# SANA: Sentiment analysis on newspapers comments in Algeria

Hichem Rahab [a,b,*], Abdelhafid Zitouni [b], Mahieddine Djoudi [c]

[a] ICOSI Laboratory, University of Khenchela, Algeria
[b] LIRE Laboratory, University of Constantine 2, Algeria
[c] TechNE Laboratory, University of Poitiers, France

A B S T R A C T

It is very current in today life to seek for tracking the people opinion from their interaction with occurring events. A very common way to do that is comments in articles published in newspapers web sites dealing with contemporary events. Sentiment analysis or opinion mining is an emergent field who's the purpose is finding the behind phenomenon masked in opinionated texts. We are interested in our work by comments in Algerian newspaper websites. For this end, two corpora were used; SANA and OCA. SANA corpus is created by collection of comments from three Algerian newspapers, and annotated by two Algerian Arabic native speakers, while OCA is a freely available corpus for sentiment analysis. For the classification we adopt Supports vector machines, naïve Bayes and k-nearest neighbors. Obtained results are very promising and show the different effects of stemming in such domain, also k-nearest neighbors gives important improvement comparing to other classifiers unlike similar works where SVM is the most dominant. From this study we observe the importance of dedicated resources and methods the newspaper comments sentiment analysis which we look forward in future works.

**Contents**

* Corresponding author at: Laboratoire ICOSI, Faculté des Sciences et de la Technologie, Bloc D, Campus Route Oum El Bouaghi, Université de Khenchela, Khenchela 40000, Algérie.
*E-mail addresses:* rahab.hichem@univ-khenchela.dz (H. Rahab), Abdelhafid.zitouni@univ-constantine2.dz (A. Zitouni), mahieddine.djoudi@univ-poitiers.fr (M. Djoudi).

## 1. Introduction

With the development of the web and its offered services, a huge amount of data is generated (Liu, 2012) and additional needs emerge to take benefit from this information thesaurus. Opinion mining from Political, economic and social data, is a new need to make the huge amount of available information in an easily understood form to decision makers in dedicated centers. Sentiment analysis vocation is to classify people opinions into specific categories to facilitate understanding the behind phenomenon.

A variety of classification approaches are available, some works deal only with positive vs. negatives classes (Rushdi-Saleh et al., 2011; Atia and Shaalan, 2015; Rahab et al., 2018), others deal with more important number of classes (Cherif et al., 2015; Ziani et al., 2013).

A very important amount of useful information is available in the comments of newspapers websites visitors around the world and in different languages. A lot of works in this era deal with English, and other European languages, but works treating Arabic language still in their beginning (Alotaibi and Anderson, 2016).

Arabic is a Semitic language spoken by about 300 million of people in 22 Arab countries. And the importance of Arabic is also that it is the language of the holy Quran (Cherif et al., 2015) the book of 1.5 billion Muslim in the world. We can find three forms of Arabic language, Classical Arabic, Modern Standard Arabic, and Dialectal Arabic. Classical Arabic is the original form of the language preserved from centuries by the Islamic literature and especially the holy Quran. For Modern Standard Arabic, it takes the role of the official language in almost all Arabic administrations. The effective spoken languages in daily conversations are Arabic dialects, which are spoken languages without a standardized writing form. They can be classified into: Levantine (spoken in Palestine, Jordan, Syrian and Lebanon) Egyptian (in Egypt and Sudan), Maghrebi (spoken in the Arab Maghreb) and Iraqi (Jarrar et al., 2017), this later one may be also divided into Iraqi versus Gulf classes (Zaidan and Callison-burch, 2011).

In these Dialect families, we will find also sub-families. In the case of the Algerian dialect, the work of (Harrat et al., 2016) classify Algerian dialects in 4 groups: 1) the dialect of Algiers and its outskirts, 2) the dialect of the east in Annaba and its outskirts, 3) the dialect Oran and the west of Algeria, and 4) the dialect of the Algerian Sahara.

Even the newspaper content is written in MSA and comments follow generally this style, we find some visitors that use Algerian Dialects words in their comments. For example the Arabic sentenceأشياء كهذه تحدث فقط في الدول المتخلفة *Âaš.yAÂkahaðihi taH.duΘu faqat fi Alduwal almutaxalifa*[1] (things like this occur only in retarded countries) is written in a comment اشياء كيما هاذي تصرى غير في الدول المتخلفة *Aaš.yA kima hAðy tas.ra γir fy Ald ∼ uwal almutaxalifa.*

Also, we found in several cases the use of د *d*, instead of ذ *ð*, which is a characteristic of the Dialect of Algiers the capital of Algeria (Harrat et al., 2016), as the case in the comment شكرا يا حفيظ هدا هو حال المسؤل الدى اسندة له المهمة و فشل اصبح ينتقد من اجل ان ينتقد و يطبل و يدافع عن الزمن الدى مر ولكن هناك رجال يصنعون المجد بتحديهم الوقائع *šukrã ya HafiyĎ hada huwa HaAl Almasŵul Al ∼ di Âus.nidat. lahu Almuhimaħ wa fašil. Asbaha yantaqid* min *Âjl Ân yantaqid wa yuTab ∼ il wa yudafiς ςan Alz ∼ aman Al ∼ adi mar wa lakin hunaka rijAl yaSnaςun Almajd bitaHad ∼ iyhim AlwaqaAĂς* (Thank you hafid this is the state of the responsible to whom is affected a mission, and he fails, so he become critic for critic, and he defends the earlier time but there are men making the glory by confronting the realities).

We are interested by comments in the Arabic Algerian online press, in the goal of developing an approach to classify these comments into positive and negative classes.

The paper is organized as follows. In the Section 2 a background of adopted methodology and used parameters are given. In the Section 3, a literature review is presented. Section 4 is dedicated to the proposed approach. An experimental study is explained and obtained results are in the Section 5. In Section 6 the achieved results are discussed. We finish by conclusion and perspectives to future works.

## 2. Background

### 2.1. Matter approach

MATTER is a cyclic approach for natural language texts annotation, the approach is based on several iterations to achieve the annotation process (Pustejovsky and Stubbs, 2012). The MATTER approach consists on a cycle of six steps. The model of the phenomenon may be revised for further train and test steps (Ide and Pustejovsky, 2017).

*Model*: in the first step the studied phenomenon will be modeled.
*Annotate*: an annotation can be seen as a metadata (Matthew and Jessica, 2010). This metadata will be added to our corpus for data classification into predefined classes like positive, negative, neutral, etc. The annotation may be integrated in the document to annotate, in a manner, that when the document is moved, the metadata still integrated, for example the addition of a distinction word in the file name. It can also take the form of a folder in which the data files are grouped, in this case a file extracted out of this folder will lose this metadata (Matthew and Jessica, 2010).

The annotation can be done at several levels.

- *Document level:* the whole document take the same label, such as: positive/negative (Rushdi-Saleh et al., 2011) or subjective/objective, ...etc.
- *Sentence level:* in this level each sentence in the document may have an independent tag, an example of this level is the tweet's classification (Brahimi et al., 2016) that the tweet cannot exceed 140 words.
- *Word level*: Also known as Part Of speech tagging POS (Tunga, 2010), where each word is tagged according to its position in the text (e.g. noun, verb, and pronoun) (Jarrar et al., 2017).

[1] For transliteration we follow in this work the scheme developed by Habash et al. (2007).

We can find several ways to achieve annotation with. Annotation by 2–5 persons having some specified skills (Alotaibi and Anderson, 2016) (Pustejovsky and Stubbs, 2012), Crowdsourcing where the annotation is done by an important number of annotators without specific skills (Bougrine et al., 2017), or Annotation based on rating systems offered by opinion sites (Rushdi-Saleh et al., 2011).

The final version of the annotated data called the gold standard is the corpus to be used in the classification step (Pustejovsky and Stubbs, 2012).

Train: a part of the data with their true classes is used to train the classifier.
Test: the rest of data (which is not used for training) is submitted to classifier for test.
Evaluate: evaluation metrics are calculated, to measure the annotation and classification performances.
Revise: based on evaluation metrics the model may be revised, and additional iteration is to do if needed.

### 2.2. Validation method

In the scope of this work the 10-fold Cross-validation method is used. Cross-Validation is, in machine learning, a method whose objective is to evaluate and compare learning algorithms. It consists of dividing the data in two segments: The first segment is used to learn or train a model and the second one is used to validate the model. In the 10-fold cross validation the corpus is divided into 10 segments of the same size, so in each iteration, 9 segments are used to train the model while the 10th is reported to the test step, this operation will be repeated in a manner that each segment is used both in the train and in the test of the model (Refaeilzadeh et al., 2009). The performance values are taken as a combination of the k performance values (as an average or another combination) to have a single estimation (Mountassir et al., 2013). The authors in (Kohavi, 1995) and (Steven and G, 1997) conclude that 10-fold cross validation is the best alternative to follow in classification process, even if computation power allows more folds.

### 2.3. Classifiers

Three well-known classifiers are used:

*Support-vector machines:* support-vector machines SVM is a relatively new machine learning method for binary classification problems (Cortes and Vapnik, 1995). To have the best results with SVM, the practitioner needs to well choice and fixed certain parameters: used kernel, gamma, and also well data collecting and pre-processing (Ben-Hur and Weston, 2010).
*Naive Bayes*: the well-known Naïve Bayes classifier is based on the "Bayes assumption" in which the document is assigned to the class in which it belongs with the highest probability (McCallum and Nigam, 1998).
*K-nearest neighbors*: k-nearest neighbors KNN is a simple classifier that use an historical values search to find the future ones (Wang, 2015).

### 2.4. Evaluation measures

1. *Inter Annotators Agreement*: several metrics are used in literature to evaluate the Inter Annotators Agreement (IAA). The kappa coefficient (Jean, 1996a) is the most used in two annotators based works (Alotaibi and Anderson, 2016; Pustejovsky and Stubbs, 2012). The coefficient is defined as:

**Table 1**
Interpretation of k parameter.

| K | Agreement level |
|---|---|
| < 0 | Poor |
| 0.01–0.20 | Slight |
| 0.21–0.40 | Fair |
| 0.41–0.60 | Moderate |
| 0.61–0.80 | Substantial |
| 0.81–1.00 | Perfect |

**Table 2**
Confusion matrix.

| | True class | |
|---|---|---|
| Predictive class | Positive | Negative |
| Positive | True positive (TP) | False Positive (FP) |
| Negative | False Negative (FN) | True Negative (TN) |

$$k = \frac{\Pr(a) - \Pr(e)}{1 - \Pr(e)}$$

where, Pr (a) represent the proportion of the cases where both annotators agree, and Pr(e) is the proportion we search that the two annotators agree by chance (Jean, 1996b). Table 1 gives a proposed interpretation of k parameter (Pustejovsky and Stubbs, 2012).

2. *Confusion matrix*: confusion matrix or contingency table is a shown in Table 2, Where:
   - o TP counts the correctly assigned comments to the positive category.
   - o FP counts the incorrectly assigned comments to the positive category.
   - o FN counts the incorrectly rejected comments from the positive category.
   - o TN counts the correctly rejected comments from the positive category.
3. *Precision and Recall*: three performance parameters were used, precision, recall, and accuracy.

$$Precision = \frac{TP}{TP + FP}$$

$$Recall = \frac{TP}{TP + FN}$$

4. *Accuracy*: precision and recall are both complementary one to the other; we combine the two using the Accuracy measure given as:

$$Accuracy = \frac{TP + TN}{TP + FP + TN + FN}$$

## 3. Related works

Sentiment Analysis is an emergent and challenging field of Data Mining and Natural Language Processing (NLP); it is a research issue with the purpose of extract meaningful knowledge from user-generated content, for tracking the mood of people about events, products or topics (G and Chandrasekaran, 2012). It may be considered as a classification problem, where the goal is to determine whether a written document, e.g. comments and reviews, express a positive or negative opinion about specific entities (Korayem et al., 2016), (Alotaibi and Anderson, 2016). It consists generally of three main steps: pre-processing, feature selection and sentiment classification (Assiri et al., 2015).

In (Rahab et al., 2017) the authors have created ARAACOM, ARAbic Algerian Corpus for Opinion Mining, 92 comments were collected from an Algerian Arabic newspaper website. Support vector machines and Naïve Bayes classifiers were used. Both uni-gram and bi-gram word model were tested. The best results are obtained in term of precision and bi-gram model increase results in almost all cases.

The authors of Curras (Jarrar et al., 2017) investigate in a corpus creation for Palestinian Arabic dialect. Two annotators are solicited to annotate morphologically Curras at the word level, and Inter Annotators Agreement is calculated using Kappa coefficient. After annotation the two annotators work together to agree in the resultant gold standard. The best accuracy among the annotators achieves 98.8%.

The work of (Abdul-Mageed and Diab, 2012) presents a multi genre corpus for Modern Standard Arabic, annotated at the sentence level. Several annotation methods were adopted, and kappa (k) parameter is used to measure inter annotators agreement (IAA). The authors conclude that a training of annotators is necessary to have a consistent annotation.

A corpus dedicated to Arabic sentiment analysis is created from tweets in (Gamal et al., 2019), the tweets are annotated (labelled) manually. Five classification algorithms are used, Support idge Regression (RR), Vector Machines (SVM), Naive Bayes (NB), Adaptive Boosting (AdaBoost), and Maximum Entropy (ME), and the best accuracy is obtained when using RR.

In (Rushdi-Saleh et al., 2011) the authors create OCA an opinion mining corpus for Arabic with 250 positive documents and 250 negative ones. The corpus is annotated at the document level by using web sites rating systems. Support vector machines and Naïve Bayes classifiers were used for evaluation. The corpus documents are mostly related to movie reviews.

The OCA corpus is used in addition to an inhouse prepared corpus in (Duwairi and El-orfali, 2013) in their study of the preprocessing effects on sentiment analysis for arabic language. SVM, NB an KNN classifiers are used, and they prove the effect of preprocessing in improving classification performance.

In their work (Tripathy et al., 2017) the authors adopt sentiment analysis at the document level. To evolve their accuracy they used SVM for feature selection and another classification method, Artificial neural network (ANN), for sentiment classification at document level. The authors have used IMDb and polarity movie reviewer datasets, and 10 cross-validation method adopted for classification. The obtained results are positively influenced by the number of hidden layers of ANN.

In (Ziani et al., 2019) a combination of Support Vector Machines and Random Sub Space algorithms is compared with an hybrid approach where the Genetic Algorithms are adopted for feature selection. The used data set is 1000 reviews collected from two Algerian newspapers and manually annotated by an expert without detailing the annotation process. It is proved that the hybrid approach can improve classification results.

From this review of literature in opinion mining works and especially works dealing with Arabic language, see Table 3, we can conclude that an important part of work concern movie reviews. So conducting studies with other topics require developing dedicated benchmarks that can be used to validate or revise existing results. Also, publicly available corpora are very sparse which make very necessary the development of dedicated resources to carry out studies is this language.

## 4. Proposed approach

In our research we adopt supervised learning, or corpus based approach for opinion mining or sentiment analysis in Arabic reviews. In this work we have used SANA our proper corpus, in addition to a well known and publically available corpus OCA[2] dedicated for Arabic sentiment analysis.

For SANA corpus creation we follow a web search in three Algerian Arabic newspaper web sites, in occurrence Echorouk[3], Elkhabar[4], and Ennahar[5]. We select articles covering several subjects (news, political, religion, sports, and society). The created corpus is available online[6].

In this work MATTER approach (Pustejovsky and Stubbs, 2012) for comments annotation is enhanced. We add a processing (PROCESS) step to have MApTTER approach. This allows us to give comments in the brute form to our annotators. So the processing step is included to the approach to:

- The annotators deal with the original text.
- The new examples can be added to any iteration.

The following algorithm summarizes our proposed approach:
Algorithm 1: Our proposed approach

***Algorithm:*** *Enhanced ARAACOM*

```
Begin
  IAA = 0;
while (IAA <= 100% ) do
  read (URL);
  Page = load (URL);
    while (there is comments in Page) do
  Extract the following Comment
  if (Comment in Data_base) then
    Delete Comment;
  Else
    Add Comment to the Data_base;
  end if
end while
MODEL
ANNOTATE
Calculate New_IAA //the New IAA
  if New_IAA <= IAA then
    go to MODEL
  end if
PROCESS
TRAIN And TEST
EVALUATE
  if (insufficient results)
    Break;
  end if
REVISE
end while
End
```

### 4.1. Model

The model is defined as the triplet: M= {T,R,I}
T = {Comment_classe, Positive, Negative, Neutral}
R = {Comment_classe::= Positive|Negative| Neutral}
I = {Positive: “Subjective with positive sentiment”,
Negative: “Subjective with negative sentiment”,
Neutral: “out of topic or without sentiment (objective)”}

[2] http://sinai.ujaen.es/wiki/index.php/OCA_Corpus_(English_version).
[3] www.echoroukonline.com/ara/.
[4] www.elkhabar.com.
[5] www.ennaharonline.com.
[6] http://rahab.e-monsite.com/medias/files/corpus.rar.

**Table 3**
Comparison of Related works approaches.

| Work | Classifiers/ methods | Dataset | Level / Classes | Best Results |
|---|---|---|---|---|
| Rahab et al. (2018) | SVM<br>NB | SIAAC (32 positive reviews and 60 negative reviews). | Document<br>Pos vs. Neg | 88.31% Of F_mesure |
| Rushdi-Saleh et al. (2011) | SVM<br>NB | OCA (250 positive reviews and 250 negative reviews). | Document<br>Pos vs. Neg | 95.20% Of Recall |
| Jarrar et al. (2017) | IAA | Curras | Word | 99.3% of accuracy |
| Abdul-Mageed and Diab (2012) | IAA | AWATIF (Penn Arabic Treebank, Wikipedia Talk Pages and Web forums) | Sentence | Kappa = 0.820 |
| Gamal et al. (2019) | NB<br>AdaBoost<br>SVM<br>RR<br>ME | 151.500 (75.744 positive and 75.744 negative) | Document (tweet)<br>Pos vs. Neg | 99.90% of Accuracy |
| Duwairi and El-orfali (2013) | SVM<br>NB<br>KNN | OCA politics dataset of 322 manually by collecting reviewers opinions from Aljazeera website | Document | 99.6% of Accuracy in OCA dataset 85.70% of Accuracy in politics dataset |
| Ziani et al. (2019) | RSS<br>SVM<br>GA | 1000 reviews from Algerian press | document | 85.99% of Accuracy |
| Tripathy et al. (2017) | SVM | IMDb (Internet movie database 12,500 positive and 12,500 negative reviews)<br>polarity dataset (1000 positive and 1000 negative reviews) | Word/ Document | 96.40% of Accuracy |

**Table 4**
Sample of French comments manual pre-processing.

| Original French comment | Translated comment in Arabic | English meaning |
|---|---|---|
| Merci Mr Hafid vous avez bien résumé qu est ce qui ce passe dans notre football | شكرا حفيظ لقد لخصت جيدا ما يحدث في كرة القدم عندنا | Thank you hafid you resume well what happen in our football |
| Grand Merci | شكرا جزيلا | Thank you very much |

In the following DTD the annotation tags and attributes were defined, to have an XML format of comments and annotation:

```
<!ELEMENT comment (#PCDATA)>
<!ATTLIST comment Sentiment (Positive | Negative | Neutral)
  #IMPLIED >
<!ELEMENT named_entity (#PCDATA)>
<!ATTLIST named_entity role (political_personality | location |
  events | organization)>
```

## 4.2. Annotation

Two Arabic native speakers are requested to annotate our corpus. In the beginning of each annotation round, a set of guidelines were given to annotators to have the best degree of contingency in obtained results.

*Annotation Guidelines*: Guidelines are orientations we give to annotators to have homogeneous annotation results. In the guidelines, the project must be described with its methodology, outcomes and all information needed to achieve our goals (Ide and Pustejovsky, 2017). In each round of the MApTTER cycle, annotation guidelines will be refined taking into account previous results.

*Adjudication*: In adjudication the annotation from different annotators are merged to have a single corpus called gold standard (Ide and Pustejovsky, 2017).

**Table 5**
Sample of Arabizi comments.

| Arabizi comment | Arabic equivalence | English meaning |
|---|---|---|
| YA3TIK SAHA KHOYA BARATLI KHATRI FI L3ADYAN VIVE L'ALGÉRIE 1.2.3 | يعطيك الصحة أخي أثلجت خاطري في الأعداء، تحيا الجزائر | Thank you brother you Warmed my heart in the enemies, life to Algeria |
| kem anta kabir ya hafidh | كم أنت كبير يا حفيظ | You are great hafid |
| BARAKA ALALHO FIK | بارك الله فيك | God bless you |

## 4.3. Processing

To have the best results in stemming and optimizing the word vector, a set of pre-processing steps are conducted:

1. *Manual text pre-processing*: We found a lot of spelling mistakes in collected comments, also some comments are written in languages other than MSA, such us French and English. First, all comments are translated into Modern Standard Arabic (MSA); we give as samples the comments in Table 4.

Second, Repeated letters such as اليوممممممممممممممممممممممم Alyaw.mmmmmmmmmmmmmmmmmmmmm (today with the last letter repeated) become اليوم Alyaw.m (today). and بعييييييييييييييييييدا baςiiiiiiiiiiiiiiiiiiiiiiydã (far with a middle letter repeated) which become بعيدا baςiydã (far).

Then Arabizi comments are transformed into their Arabic equivalent as shown in Table 5. Arabizi is an Arabic language used in SMS and tchat on the Internet, it differs from transliteration that there is no standard to adopt in this language.

We finish by character encoding where all texts are resolved to UTF-8 encoding format.

2. *Tokenization*: In tokenization, words are separated by non-letters characters.
3. *Stemming*: light stemming is used in this step. Figs. 1 and 2 show light stemming and stemming of the same comment. We remark that the word الجزائر *AljazAŷir*(Algeria with the definite article) for example in light stemming is stemmed to جزائر *jazAŷir* (Algeria without the definite article), when in stemming is stemmed as جزر *jazar* or *jaz.r* (carrot or ebb). And the word اقتصاديا *Ăq.tiSAdiyã* (economically) in light stemming is stemmed as it without changing اقتصاديا *Ăq.tiSAdiyã*, when in stemming it is stemmed as قصد *qaSada* (intention or meaning). In the scope of this work we have used light stemming, because stemming generate the root of the word which gives a different meaning in several cases (Mountassir et al., 2013).

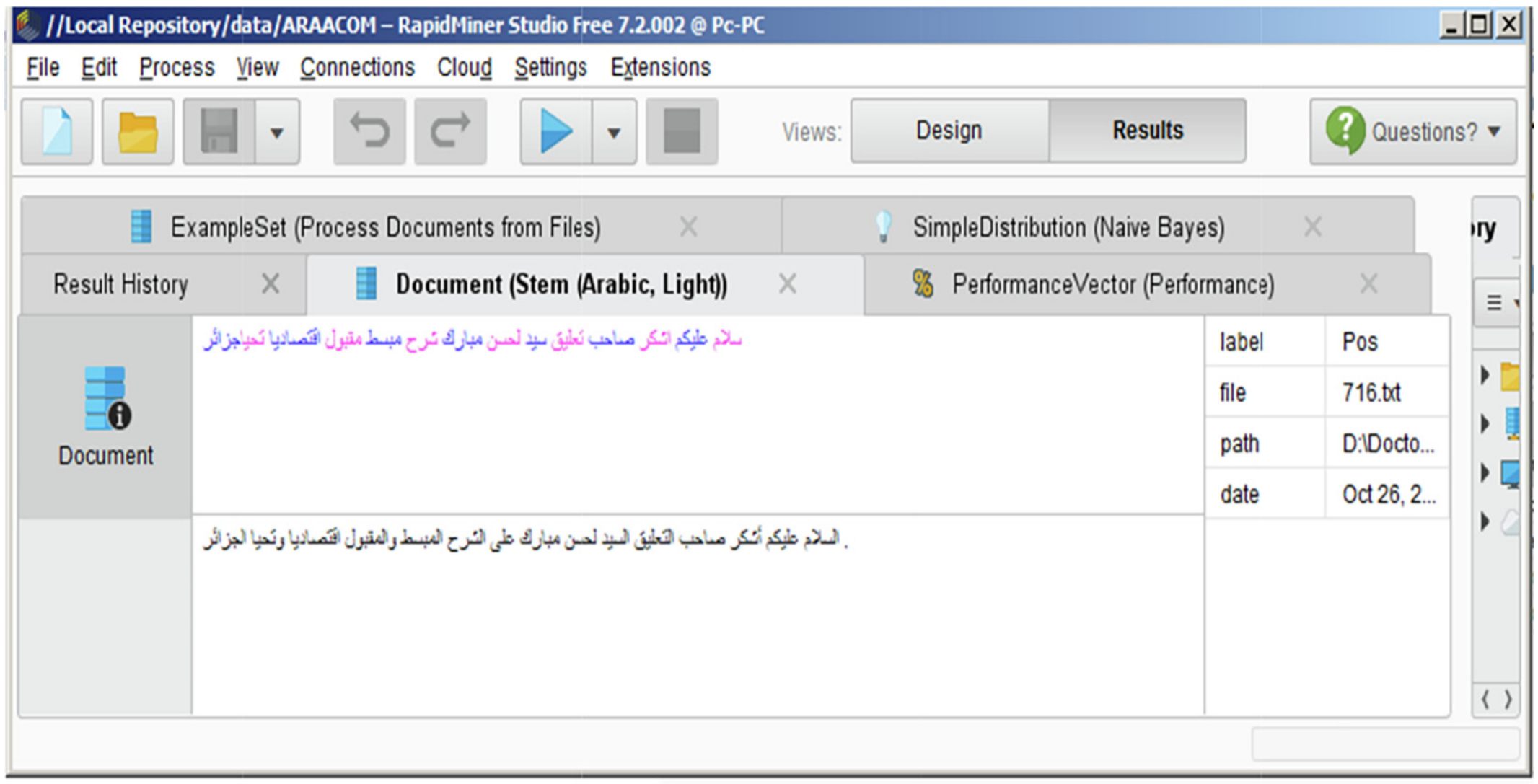

**Fig. 1.** Sample of word light stemming.

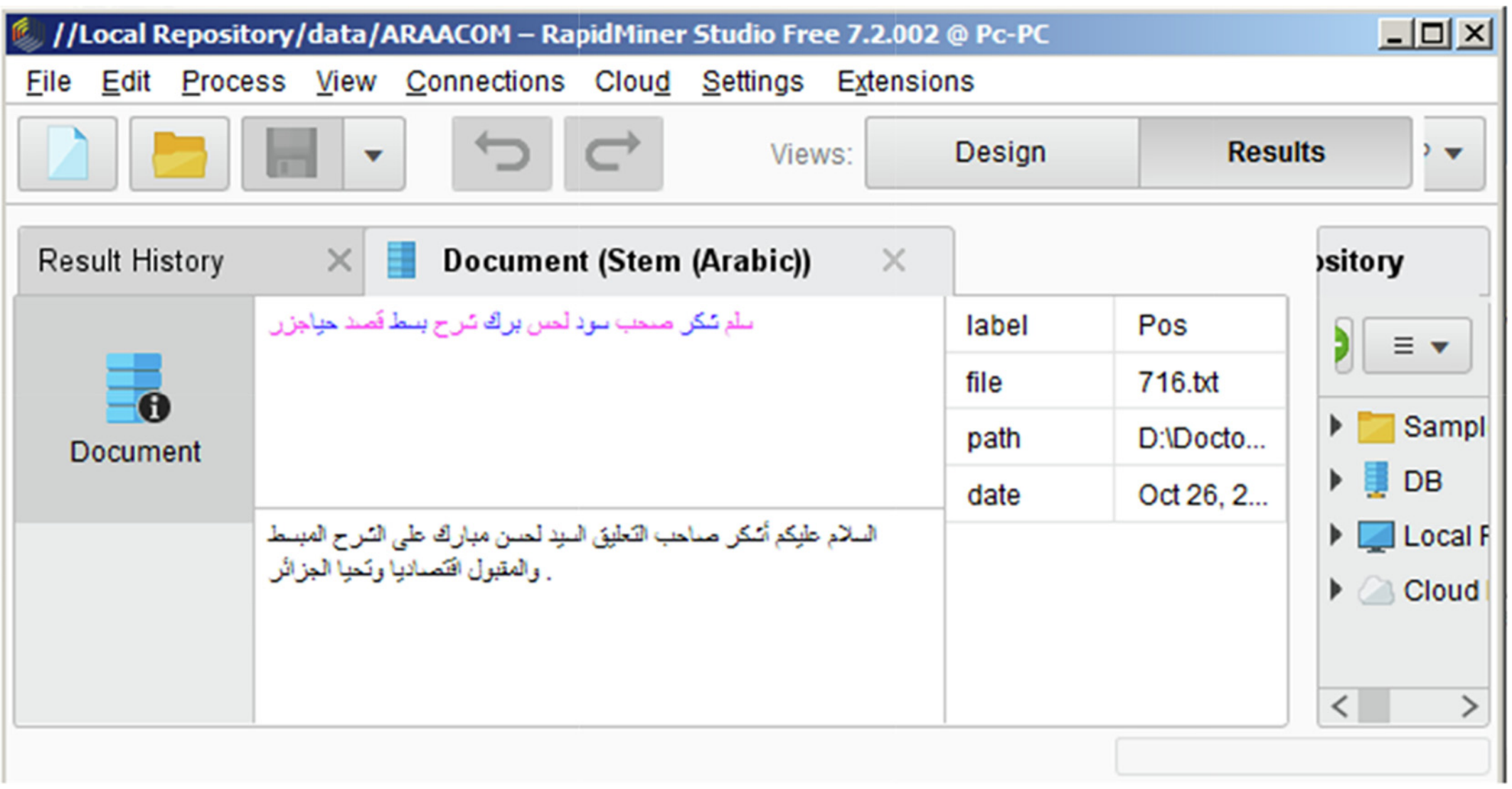

**Fig. 2.** Sample of word Stemming.

4. *Stop words Removal*: a list of stop words is offered by the used toolkit.
5. *Word n-gram*: Uni-gram, bi-gram and tri-gram word are generated.
6. *Word vector*: the four vectors are tested (Term frequency TF, Term occurrence TO, Term frequency Inverse document frequency TF-IDF and Binary term occurrence BTO).

### 4.4. Train and test

In this step three classification methods, Support vector machines (SVM), Naïve Bayes (NB), and k-nearest neighbors (KNN) were used. And the 10-fold cross validation is adopted. For the SVM we use SVM linear, and for KNN we adopt the k = 9 parameter as suggested in Brahimi et al. (2016).

### 4.5. Evaluate

For evaluation, Precision Recall and Accuracy are calculated for each classifier and obtained resultants were compared and discussed in each round of the MApTTER cycle.

### 4.6. Revise

In revision step adopted approach is revised in the light of evaluation metrics, and a decision to continue or stop the MApTTER cycle is to take in each step.

## 5. Experimental study

### 5.1. First round

1. *Model*: The above model is conserved as it. M={T,R,I}

**Table 6**
Confusion matrix of the first round.

| | | Annotator 02 | | | Total |
|---|---|---|---|---|---|
| | | Positive | Negative | Neutral | |
| Annotator 01 | Positive | 34 | 2 | 6 | 42 |
| | Negative | 10 | 65 | 21 | 96 |
| | Neutral | 10 | 6 | 24 | 40 |
| Total | | 54 | 73 | 51 | 178 |

**Table 7**
First round Gold Standard.

| Positive | Negative | Neutral | Total |
|---|---|---|---|
| 45 | 88 | 45 | 178 |

T = {Comment_classe, Positive, Negative, Neutral}
R = {Comment_classe::= Positive| Negative| Neutral}
I = {Positive: "Subjective with positive sentiment", Negative: "Subjective with negative sentiment", Neutral: "out of topic or without sentiment (objective)"}.

2. *Annotate*: The Annotation guidelines given to annotators are, to annotate a comment as it presents a positive, negative or neutral sentiment regarding the article topic, so for each comment the annotators have the correspondent article, Table 6.
   - o *IAA Inter Annotators Agreement:*

$$k = \frac{0.6910 - 0.3569}{1 - 0.3569} = \frac{0.3341}{0.6431}$$

So: $k = 0.5195$ considered as moderate (see Table 1)

   - o *Adjudication*: In adjudication step the two annotators are working together in the goal of obtaining a consensus in annotation and in cases when a consensus was not obtained the comment is considered as neutral. So we reach the following gold standard (Table 7).

We take in this work only the positive and negative classes, to have equilibrium we generate the corpus with the 45 positive comments and 45 of the negative ones.

3. *Processing*: The processing steps describing above are doing again, namely: text pre-processing, UTF-8 encoding, tokenization, stemming, stop words removal, n-gram word generation, and word vectors creation.
4. *Train and Test*: For these two steps cross validation method was adopted using the three classifiers, support vector machines, naïve Bayes and k-nearest neighbors.
5. *Evaluate*: To evaluate our model we calculate precision and recall of both negative and positive classes and accuracy of the classification at whole.

In Table 8 are presented evaluations of first round classification of SANA corpus, word weighting has different impacts according to classification method, and whether the light stemming is used. The SVM method gives best performances using TO and BTO word weighting without stemming, while one using the light stemming, are TF and TF-IDF word weighting the best performing. In the case of NB classifier we have best results using TO and TF-IDF whatever the light stemming is adopted or not. For the KNN the best results are obtained using TF and TF-IDF word weighting.

Also, the use of light stemming increases the classification performance in almost cases for the three classification methods.

6. Revise: After evaluation of the IAA and classification scores some suggestions are mentioned to take into account in the next round:
   - o In the annotation guidelines the authors must take into account only the content of the comment no matter what is the article subject. So this recommendation may conduct to have more consensus between annotators, that is in the previous round, when taking into account the article topic, we observe that for the same idea (comment) one annotator can have a positive sentiment and the other have a neutral or even negative sentiment.

### 5.2. Second round

*Model*: We conserve the above model without any changes.
*Annotate*: In this round we give our annotators instruction to take into account only the content of the comment, as it is recommended in the revision of the above step (see Table 9).
o IAA Inter Annotators Agreement:

$$k = \frac{0.7212 - 0.3333}{1 - 0.333} = \frac{0.3882}{0.667}$$

$k = 0.5820$

The Inter Annotator Agreement is improved with the new instruction, so it achieves k = 0.5820.

o *Adjudication:* We generate our gold standard by adjudication of annotation step. And this is performed to obtain a single annotated corpus from the two available.

We eliminate the neutral class and for equilibrium we take the 194 negative comments and only 194 from the positive comments (Table 10).

*Processing*: The above described processing steps are conducted twice.
*Train and Test*: We always use the three classifiers, SVM, NB and KNN. And the 10-fold cross-validation method.
Evaluate: In the Table 11 we show the accuracy of the second round classification of SANA corpus. As in the first round we remark that word weighting has different impacts according to the using algorithm.

Also, bi-gram and tri-gram models effect is very weak and give low differences in performance results.

*Revise*: In this point we decide to stop this cycle and report the obtained results.

### 5.3. OCA corpus

In the case of OCA corpus (Table 12) best performance results are obtained using word vector weighting TF and TF-IDF both in SVM and KNN classifiers. And the stemming increase accuracy in both methods.

**Table 8**
First round accuracy of SANA corpus.

| Light Stem | | SVM | | | NB | | | KNN | | |
|---|---|---|---|---|---|---|---|---|---|---|
| | | Unigram | Bigram | Tri-gram | Unigram | Bigram | Tri-gram | Unigram | Bigram | Tri-gram |
| No | TO | **58.89** | **58.89** | **57.78** | **64.44** | **61.11** | **62.22** | 52.22 | 51.11 | 51.11 |
| | TF | 55.56 | 54.44 | 54.44 | 63.33 | 58.89 | 60.00 | **57.78** | **56.67** | **56.67** |
| | TF-IDF | 52.22 | 52.22 | 53.33 | **65.56** | **63.33** | **63.33** | **63.33** | **63.33** | **61.11** |
| | BTO | **60.00** | **58.89** | **57.78** | 58.89 | 57.78 | 58.89 | 50.00 | 52.22 | 52.22 |
| Yes | TO | 63.33 | 58.89 | 60.00 | **68.89** | **63.33** | **64.44** | 48.89 | 51.11 | 52.22 |
| | TF | **70.00** | **65.56** | **66.67** | 66.67 | 63.33 | 63.33 | **70.00** | **68.89** | **68.89** |
| | TF-IDF | **67.78** | **67.78** | **68.89** | **70.00** | **65.56** | **66.67** | **64.44** | **64.44** | **64.44** |
| | BTO | 62.22 | 62.22 | 63.33 | 66.67 | 61.11 | 62.22 | 53.33 | 51.11 | 51.11 |

**Table 9**
Second round confusion matrix.

| | | Annotator 02 | | | Total |
|---|---|---|---|---|---|
| | | Positive | Negative | Neutral | |
| Annotator 1 | Positive | 161 | 08 | 11 | 180 |
| | Negative | 34 | 94 | 50 | 178 |
| | Neutral | 14 | 26 | 115 | 155 |
| Total | | 209 | 128 | 176 | 513 |

**Table 10**
Second round Gold Standard.

| Positive | Negative | Neutral | Total |
|---|---|---|---|
| 236 | 194 | 83 | 513 |

When for NB, the best results are obtaining with TO and BTO word weighting. The stemming in the case of NB gives low performance in almost cases.

## 6. Results discussion

The obtaining results differ from a corpus to another, and this is due to nature of the corpus. In the case of OCA corpus which is constituted from movie reviews, and written in well structured MSA. The best results for KNN and SVM are improved with TF and TF-IDF, that this two word weighting methods considers the word in the context of its document and the corpus at a whole. While in the case of NB classifier, the best results are obtained in TO and BTO, that consider the weight of a term independently of the context, and NB as classifier consider the independence between features given the context of the class. The light stemmer does not give an improvement in this case. The bi-gram and tri-gram mod-els give a considerable amelioration; we report this improvement to result of negation and regular expression used in well written and homogenous documents.

**Table 11**
Second round accuracy of SANA corpus.

| Light Stem | | SVM | | | NB | | | KNN | | |
|---|---|---|---|---|---|---|---|---|---|---|
| | | Unigram | Bigram | Tri-gram | Unigram | Bigram | Tri-gram | Unigram | Bigram | Tri-gram |
| No | TO | **71.13** | **69.85** | **71.13** | 70.36 | 70.62 | 70.36 | **64.69** | **63.92** | **63.66** |
| | TF | 70.62 | 70.88 | 70.62 | 70.36 | 71.13 | 71.13 | 61.60 | 61.86 | 62.11 |
| | TF-IDF | 69.33 | 69.07 | 70.88 | **70.36** | **71.91** | **72.16** | 56.19 | 56.96 | 57.73 |
| | BTO | **71.39** | **70.10** | **68.56** | **72.16** | **71.65** | **70.88** | **63.92** | **62.89** | **62.37** |
| Yes | TO | **71.13** | 69.33 | **71.91** | 73.45 | 72.94 | 72.42 | 64.69 | 65.72 | 65.72 |
| | TF | 70.88 | **72.16** | 71.39 | **73.97** | **75.00** | **74.74** | **65.72** | **67.78** | **67.53** |
| | TF-IDF | 69.33 | 72.16 | 69.85 | 73.45 | 73.97 | 73.97 | **65.21** | **66.49** | **66.49** |
| | BTO | **71.65** | **69.33** | **72.16** | **74.74** | **74.23** | **73.71** | 63.66 | 62.63 | 62.63 |

**Table 12**
Accuracy of OCA Corpus.

| Light Stem | | SVM | | | NB | | | KNN | | |
|---|---|---|---|---|---|---|---|---|---|---|
| | | Unigram | Bigram | Tri-gram | Unigram | Bigram | Tri-gram | Unigram | Bigram | Tri-gram |
| No | TO | 70.00 | 70.80 | 71.00 | **86.00** | **89,00** | **89,80** | 51.40 | 51.60 | 51.60 |
| | TF | **77.00** | **80.80** | **80.00** | 85.20 | 87.80 | 88.00 | **88.00** | **88.00** | **88.00** |
| | TF-IDF | **77.80** | **80.80** | **80.40** | 85.60 | 87.00 | 87.20 | **87.80** | **88.40** | **88.40** |
| | BTO | 72.20 | 72.80 | 73.40 | **87.40** | **88.80** | **89.20** | 50.00 | 50.00 | 50.00 |
| Yes | TO | 67.40 | 68.80 | 68.80 | **83.00** | **87.00** | **87.60** | 51.20 | 50.80 | 51.00 |
| | TF | **79.40** | **82.00** | **82.80** | 81.20 | 86.20 | 86.80 | **79.80** | **83.40** | **84.80** |
| | TF-IDF | **78.20** | **82.00** | **82.00** | 80.80 | 86.40 | 86.60 | **80.40** | **83.20** | **83.60** |
| | BTO | 69.20 | 71.40 | 70.80 | **84.40** | **89.20** | **88.40** | 50.00 | 50.00 | 50.00 |

When with the SANA (especially second round) we observe that word weighting results are different from a classifier to another. The light stemmer effects are better than OCA corpus and we suggest the light stemmer for mixture document (dialectal and MSA). Bi-gram and tri-gram improvement are very weak, so there are no commonly used expression in SANA corpus compared to OCA corpus due to the nature of each one.

## 7. Conclusion and perspectives

Available corpora on the web for carried out Arabic sentiment analysis studies are very rare, and those available are generally related to movie and film reviews due to the available comments in such websites. The case of newspaper comments is more delicate, that it is related to the country in which it is exists, so dealing with such comments must take into account, not only Arabic language but also related dialects and used languages in this country. In this work we proposed our enhanced approach for opinion mining in Algerian Arabic Newspapers comments. We use two corpora, the first is SANA which is created in the scope of this work, and the second is the well known and available in the web the OCA corpus.

For SANA annotation, MApTTER approach is used to annotate Algerian newspaper comments. MApTTER is based on an existing annotation approach MATTER.

Three classifiers were used, SVM, NB and KNN. Obtained results are promising, but still to develop. And the most important conclusion is about light stemming that give different improvement depending on the nature of the corpus, and it is suggested when dealing with non homogenous and not well written documents. Also, the bigram and trigram models are not always a good representation regarding the obtained results, their effect depend on the formulation of corpus documents.

We aim in the future to develop the approach by taking into account the parts in the comment that enclose the most semantic, which is the first and the last parts.

## Conflict of interest

None.